\documentclass[preprint,2p,times]{elsarticle}

\usepackage{amsmath}
\usepackage{natbib}
\usepackage{eurosym}
\usepackage{fullpage}
\usepackage{graphicx}
\usepackage{wrapfig}
\usepackage[english]{babel}
\usepackage[center]{caption}
\usepackage{mathtools}
\usepackage{mathptmx}
\usepackage{latexsym}
\usepackage{url}
\usepackage{multirow}
\usepackage{amssymb}
\usepackage{capt-of, blindtext}
\usepackage{pdfpages}
\usepackage{setspace}
\usepackage{lineno}
\usepackage{algorithmic}
\usepackage{algorithm}

\usepackage{geometry}
\date{}

\begin{document}

\title{Towards a Quantum-Like Cognitive Architecture for Decision-Making} 

\begin{frontmatter}
	\author[{label1}]{Catarina Moreira} \ead{catarina.pintomoreira@qut.edu.au}
	\author[label1]{Lauren Fell}\ead{l3.fell@qut.edu.au}
	\author[label1]{Shahram Dehdashti}\ead{shahram.dehdashti@qut.edu.au}
	\author[label1]{Peter Bruza}\ead{p.bruza@qut.edu.au}
	\author[label2]{Andreas Wichert} \ead{andreas.wichert@tecnico.ulisboa.pt}
	\address[label1]{School of Information Systems, Queensland University of Technology, 2 George St, Brisbane City, QLD 4000, Australia}
	\address[label2]{Instituto Superior T\'{e}cnico / INESC-ID, Av. Professor Cavaco Silva, 2744-016 Porto Salvo, Portugal}


	\begin{abstract}  
We propose an alternative and unifying framework for decision-making that, by using quantum mechanics, provides more generalised cognitive and decision models with the ability to represent more information than classical models. This framework can accommodate and predict several cognitive biases reported in Lieder \& Griffiths without heavy reliance on heuristics nor on assumptions of the computational resources of the mind.

\end{abstract} 

\end{frontmatter}

\doublespace

\section{Comment}
Lieder \& Griffiths (L\&G) propose a normative bounded resource-rational heuristic function to relax the optimality criteria of the Expected Utility Theory and justify the choices that lead to less optimal decisions. Expected utility theory and classical probabilities tell us what people should do if employing traditionally rational thought, but do not tell us what people do in reality (Machina, 2009). Under this principle, L\&G propose an architecture for cognition that can serve as an intermediary layer between Neuroscience and Computation. Whilst instances where large expenditures of cognitive resources occur are theoretically alluded to, the model primarily assumes a preference for fast, heuristic-based processing. We argue that one can go beyond heuristics and the relaxation of normative theories like the Expected Utility Theory, in order to obtain a unifying framework for decision-making. 

The proposed alternative and unifying approach is based on a quantum-like cognitive framework for decision-making, which not only has the ability to accommodate several paradoxical human decision scenarios along with more traditionally rational thought processes, but can also integrate several domains of the literature (such as Artificial Intelligence, Physics, Psychology and Neuroscience) into a single and flexible mathematical framework. 

Figure 1 presents the proposed unifying quantum-like framework for decision-making grounded in the mathematical principles of quantum mechanics without the specification of heuristics to accommodate the cognitive biases addressed by L\&G. 

\begin{figure}[h!]
\centering
\includegraphics[scale=0.4]{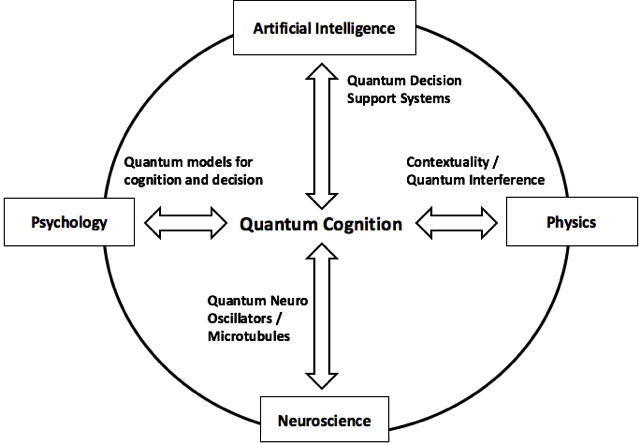}
\caption{A quantum-like cognitive architecture}
\end{figure}

In quantum cognitive models, events are represented as multidimensional vectors according to a basis in complex Hilbert spaces, which reflects the potentials of all events. In quantum mechanics, this property refers to the superposition principle. For instance, when making a judgment whether or not to buy a car, a person is at the same time in an indefinite state corresponding to buy and, in the state, to not buy (Figure 2). Each person reasons according to their own basis. Different personal beliefs are obtained by rotating their basis, leading to different representations of the decision scenario and different beliefs according to each decision-maker (Figure 3). Ultimately, a person can be in a superposition of thoughts in a wave-like structure. This can create interference effects leading to outcomes that cannot be predicted by classical theories. Interference is one of the core concepts in quantum cognition.

\begin{figure}[h!]
\parbox{.3\columnwidth}{
	\includegraphics[scale=0.42]{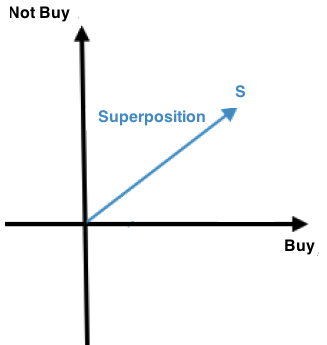}
	\caption{Hilbert space represenation of a basis state}
	\label{fig:init}
	} \hfill 
\parbox{.7\columnwidth}{
	\includegraphics[scale=0.55]{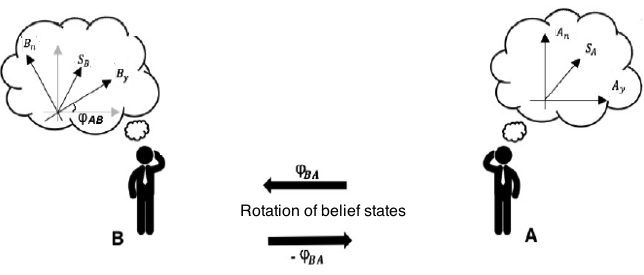}
	\caption{Each person reasons according to a N-dimensional vector space by rotating their basis state towards their personal beliefs.}
	\label{fig:end}	
	} 
\end{figure}

This wave-like paradigm enables the representation of conflicting, ambiguous and uncertain thoughts (Busemeyer and Bruza, 2012) and also undecidability (Cubitt et al., 2015). The superposition vector representation obeys neither the distributive axiom of Boolean logic nor the law of total probability. As a consequence, it enables the construction of more general models that cannot be captured in traditional classical models. The accessibility of information in the quantum cognitive framework is much higher than in a classical system, making it possible to model the different minds of bounded cognitive resources. This additional information can also accommodate several paradoxical decision scenarios, generate novel non-reductive models of how humans process concepts and generate new understandings of human cognition (Frauchiger and Renner, 2018, Vourdas, 2019). These distinctive features of quantum theory provide several advantages and more accurate and elegant explanations for empirical data in situations where classical probability theory alone leads to puzzling and counterintuitive predictions (cognitive bias, order effects, conjunction/disjunction errors…). While classical probability traditionally assumes independence of events, quantum theory provides probabilistic inferences which are strongly context dependent: the same predicate may appear plausible or not, depending upon the decision-maker's point of view (Pothos and Busemeyer, 2013). 

L\&G make a strong assumption that the mind is a computational architecture, which uses certain classes of algorithms that make the trade-off between the computational cost of using the mind's resources (and getting the necessary information) and the utility of finding the correct solution of a cognitive problem (specified at a computational level). The proposed quantum cognitive framework imposes no such assumptions about the human mind and rests on two important aspects of quantum mechanics: contextuality and interference.

Contextuality entails the “impossibility of assigning a single random variable to represent the outcomes of the same measurement procedure in different measurement conditions” (Acacio De Barros and Oas, 2015, p. 1). As a consequence, it is not possible to define a single joint probability distribution from the empirical data collected from different measurement conditions such that the empirical data can be recovered by marginalizing the joint distribution. Recent empirical evidence suggests that contextuality manifests in cognitive information processing (Cervantes et al., 2018, Basieva et al. 2019). Should contextuality be present, then it would call into question the assumption of the distribution $P(results~|~ s_0,h,E)$ (Eq 4). The intuition here is that the cognitive agent cannot form this distribution because the functional identity of the random variable of which ‘result' is an outcome is not unique across the environments E (viewed as measurement conditions in regard to the quote above). (Dzafarov \& Kujala 2014, Dzhafarov \& Kujala 2016). Although it is theoretically speculative to associate quantum-like contextuality with (Eq 4), we do so to draw attention the fact that contextuality has little known and undiagnosed consequences for the development of probabilistic models in cognitive science. 

Interference has a major impact in cognitive models, because in decision scenarios under uncertainty, one can manipulate these quantum interference effects to disturb classical probabilities and, consequently accommodate most of the cognitive limitations reported in L\&G and provide alternative inferences under uncertainty that are not captured by classical probability (Pothos and Busemyer, 2009). Recent studies suggest that quantum interference can be used to model real-world financial scenarios with high levels of uncertainty, showing a promising approach for decision support systems and artificial intelligence (Moreira et al., 2018, Moreira \& Wichert, 2018). In Neurosicence, interference effects in the brain can occur if neuronal membrane potentials have wave-like properties (de Barros \& Suppes 2009).

To conclude, we commented on the architecture proposed by L\&G, which assumes a preference for fast, heuristic-based processing and strong computational assumptions about the human mind. We proposed a unifying framework for decision-making based on quantum mechanics that provides more generalised decision models capable of representing more information than classical models. It can accommodate several paradoxical findings and cognitive biases and lead to alternative and powerful inferences focused on the perspective-dependency of the decision-maker. 

\section{Acknowledgements}
Dr. Andreas Wichert was supported by funds through Fundação para a Ciência e Tecnologia (FCT) with reference UID/CEC/50021/2019. The funders had no role in study design, data collection and analysis, decision to publish, or preparation of the manuscript.

\section{References}
Basieva, I., Cervantes, V.H., Dzhafarov, E.N., Khrennikov, A. (in press). True contextuality beats direct influences in human decision making. Journal of Experimental Psychology: General.

Busemeyer, J. and Bruza, P. (2012) Quantum Models for Cognition and Decision, Cambridge University Press

Cervantes, V.H., and Dzhafarov, E.N. (2018). Snow Queen is evil and beautiful: Experimental evidence for probabilistic contextuality in human choices. Decision 5, 193-204.

Cubitt, Toby. S., Perez-Garcia, D. and Wolf, M. (2015). Undecidability of the spectral gap, Nature, 528, 207
 
de Barros, A. and Suppes, P. (2009) Quantum Mechanics, Interference and the Brain. Journal of Mathematical Psychology, 53, 306 – 313.

de Barros, J.A., and Oas, G. (2015). Some examples of contextuality in physics: Implications to quantum cognition. arXiv:1512.00033. 
Dzhafarov, E.N., Kujala, J.V. (2014). Contextuality is about identity of random variables. Physica Scripta T163, 014009

Dzhafarov, E.N., \& Kujala, J.V. (2016). Context-content systems of random variables: The contextuality-by-default theory. Journal of Mathematical Psychology 74, 11-33.

Evans, J. S. B. (2003). In two minds: dual-process accounts of reasoning. Trends in cognitive sciences, 7(10), 454-459.

Frauchiger, D. and Renner, R. (2018) Quantum Theory Cannot Consistently Describe the use of Itself. Nature Communications, 9.

Machina, M. (2009). Risk, ambiguity, and the rank-dependence axioms. Journal of Economic Review, 99(1), 385-392.

Moreira, C., and Wichert (2018). A., Are quantum-like Bayesian networks more powerful than classical Bayesian networks?, Journal of Mathematical Psychology, 82: 73-83

Moreira, C., Haven, E., Sozzo, S., and Wichert (2018). A., Process mining with Real World Financial Loan Applications: Improving Inference on Incomplete Event Logs, PLOS ONE, 13(12): e0207806

Pothos, E. M., and Busemeyer, J. R. (2009). A quantum probability model explanation for violations of ‘rational' decision theory. Proceedings of the Royal Society, B: Biological Sciences, 276 (1665), 2171-2178)

Pothos, E. M., and Busemeyer, J. (2013). Can quantum probability provide a new direction for cognitive modeling?  Behavioral and Brain Sciences, 36, 255-274.

Vourdas, A. (2019) Probabilistic Inequalities and Measurements in Bipartite Systems. Journal of Physics A: Mathematical and Theoretical, 52 085301

\end{document}